\pdfoutput=1

\documentclass[11pt]{article}

\usepackage{ACL2023}

\usepackage{times}
\usepackage{latexsym}
\usepackage[T1]{fontenc}
\usepackage[utf8]{inputenc}
\usepackage{microtype}
\usepackage{inconsolata}
\usepackage[ruled]{algorithm2e}

\usepackage{multirow}
\usepackage{amsmath}
\usepackage{capt-of}
\usepackage{tabularx}
\usepackage{epsfig}
\usepackage{amssymb}
\usepackage{amsfonts}
\usepackage{booktabs}
\usepackage{scalerel}
\usepackage[inline]{enumitem}
\usepackage{listings}
\usepackage{varwidth}
\usepackage[export]{adjustbox}
\usepackage{tikz}
\usetikzlibrary{tikzmark}

\usepackage{stmaryrd}
\usepackage{bbm}
\usepackage{wrapfig}
\usepackage{pifont}

\definecolor{deepblue}{rgb}{0,0,0.5}
\definecolor{officeblue}{RGB}{0,102,204}
\definecolor{deepred}{rgb}{0.6,0,0}
\definecolor{deepgreen}{rgb}{0,0.5,0}
\definecolor{mybrickred}{RGB}{182,50,28}

\definecolor{fillcolor}{RGB}{216,217,252}

\usepackage{amsmath,amsfonts,bm}

\def\eqref#1{equation~\ref{#1}}

\def\1{\bm{1}}

\def\va{{\bm{a}}}

\def\vk{{\bm{k}}}

\def\vq{{\bm{q}}}

\def\vv{{\bm{v}}}

\DeclareMathAlphabet{\mathsfit}{\encodingdefault}{\sfdefault}{m}{sl}
\SetMathAlphabet{\mathsfit}{bold}{\encodingdefault}{\sfdefault}{bx}{n}

\newcommand{\softmax}{\mathrm{softmax}}

\usepackage{pifont}

\lstdefinestyle{mystyle}{
    commentstyle=\color{codegreen},
    keywordstyle=\color{magenta},
    numberstyle=\tiny\color{codegray},
    stringstyle=\color{codepurple},
    basicstyle=\footnotesize,
    breakatwhitespace=false,         
    breaklines=true,                 
    captionpos=b,                    
    keepspaces=true,                 
    numbers=left,                    
    numbersep=5pt,                  
    showspaces=false,                
    showstringspaces=false,
    showtabs=false,                  
    tabsize=2
}
 
\newcommand\rope{\textsc{RoPE}}
\newcommand\expo{\textsc{xPos}}
\newcommand\our{\textsc{LeX}}
\newcommand\ourtrm{\textsc{LeX} Transformer}
\title{A Length-Extrapolatable Transformer}

\author{%
Yutao Sun,~Li Dong,~Barun Patra,~Shuming Ma \\ \textbf{Shaohan Huang,}~\textbf{Alon Benhaim,}~\textbf{Vishrav Chaudhary,}~\textbf{Xia Song,}~\textbf{Furu Wei}\\
Microsoft \\
\url{https://github.com/microsoft/torchscale} \\}

\date{}

\begin{document}
\maketitle
\begin{abstract}
Position modeling plays a critical role in Transformers.
In this paper, we focus on length extrapolation, i.e., training on short texts while evaluating longer sequences. We define \textit{attention resolution} as an indicator of extrapolation. Then we propose two designs to improve the above metric of Transformers. Specifically, we introduce a relative position embedding to explicitly maximize attention resolution.
Moreover, we use blockwise causal attention during inference for better resolution. We evaluate different Transformer variants with language modeling.
Experimental results show that our model achieves strong performance in both interpolation and extrapolation settings.
The code will be available at \url{https://aka.ms/LeX-Transformer}.
\end{abstract}

\section{Introduction}
\label{sec:intro}

\begin{table*}[t]
\centering
\begin{tabular}{@{}lcc}
\toprule
\bf Models           & \bf Translation Invariance & \bf Length Extrapolation \\
\midrule
\multicolumn{3}{l}{\emph{Absolute Position Modeling}}                                \\
Transformer (Sinusoidal) &\ding{56}&\ding{56\ding{56}} \\
GPT-2 (Learnable) &\ding{56}&\ding{56}\ding{56} \\
\midrule
\multicolumn{3}{l}{\emph{Relative Position Modeling}}                                \\
PaLM / Roformer (\rope{})       &\ding{52}&\ding{56} \\
T5         &\ding{52}&\ding{56} \\
BLOOM / Alibi      &\ding{52}&\ding{52} \\
\ourtrm{} (Ours)     &\ding{52}&\ding{52}\ding{52} \\
\bottomrule
\end{tabular}
\caption{Position modeling capabilities of Transformer variants for language modeling.}
\end{table*}

Transformer~\citep{transformer} shows a strong performance in NLP and becomes a universal choice nowadays~\citep{vit, clip, beit3}. However, most of them have a crucial shortcoming: they can only deal with the in-distribution size of inputs. It is usually infeasible to train a model with all possible input lengths.
Therefore, a length-extrapolatable Transformer is essential for wider usage.

In sequence modeling, position information plays a crucial role in building the correct representation and understanding of the latent meaning. For Recurrent Neural Networks such as LSTM~\citep{lstm}, the calculation is done along the sequence order in O(n) time. However, the parallel attention module makes it hard to encode position effectively. First, \citet{transformer} propose absolute sinusoidal position embedding, and \citet{bert} adjust it to a learnable one. The absolute design is computation-efficient, but not comparable with subsequent relative ones~\citep{relpos, rotary, alibi}. Among many relative position embeddings, \rope{}~\citep{rotary} shows better performance and is used to many PLMs such as PaLM~\citep{palm}. However, it can't deal with sequences with exceed length. Alibi~\citep{alibi} mitigates the extrapolation problem but sacrifices the general performance.

Since different strategies concentrate on some part of the position feature, it is essential to build a comprehensive view and guide the Transformer's design systematically. First, a Transformer should be sensitive to order. Otherwise, it will degenerate into a bag-of-word model which confuses the whole meaning. Then, position translation can't hurt the representation a lot especially combing with the proper attention-mask operations. After that, a good sequence model needs to deal with any input length. As illustrated before, the length problem is not universal but special for Transformer. Especially, when a Transformer is pre-trained under a maximal length, it is not affordable to re-train for applying to tasks with longer sequences. Finally, when a Transformer satisfies the principles above, we will evaluate the performance, which requires thorough experiments and empirical analysis.

Considering all the properties above, we propose Extrapolatable Position Embedding (\expo{}), which is a universal-good design for Transformers. Based on \rope{}'s design, we propose \textit{attention resolution} as a metric to measure position monotonicity accurately. Then, we generalize its mathematical form, where an exponential decay is added to the rotation matrix. \expo{} preserves the advantage of \rope{}, and behaves stably at long-term dependency.
%
Besides, we use blockwise causal attention to increase attention resolution, which improves the performance of length extrapolation for language modeling.


We train different Transformers from scratch. On the pre-training corpus, \ourtrm{} reaches minimal perplexity on the validation set. We use the arXiv dataset (above 6k length) to evaluate the model's ability for extrapolation length. Our methods can continue decreasing the perplexity while other methods either can't extrapolate or increase the perplexity when the input length is very long.

We summarize our contributions as follows:
\begin{itemize}
\item We summarize the design principles of Transformers for position modeling.
\item We define attention resolution to indicate length extrapolation.
\item We propose an extrapolatable position embedding and use blockwise causal attention to improve length extrapolation.
\item We conduct experiments on language modeling and show that the proposed \ourtrm{} achieves strong performance on both short and long texts.
\end{itemize}

\section{Design Principles of Transformers for Position Modeling}
\label{sec:principle}

\subsection{Order Variance}

Transformer aims to capture long-term dependency efficiently~\citep{transformer}, so the distance between every two tokens is 1. Transformer without position information is actually a bag-of-word model. With effective position information, Transformer models should be variant with permuting the order~\citep{2022positionoverview}:
\begin{equation}
    f(P_\pi(X))\neq P_\pi(f(X))
\end{equation}

Although for some tasks, bag-of-words models can achieve comparable performance~\citep{wang2020position}, position information is essential generally for sequence modeling. Almost every position modeling strategy satisfies this goal~\citep{transformer, bert, relpos, wang2020position, t5, rotary}.

\subsection{Translation Invariance}

The representation of a sequence should be robust with the position's translation. For instance, in fact, a sentence's meaning is variant with padding before or after the whole sentence. We give a general form for translation invariance similar with~\citep{wang2020position}: for a Transformer model $f(\text{input}, \text{mask})$, any input sequence $X=[x_0, x_1, ..., x_n]$ with mask $M=[m_0, m_1, ..., m_n]$, the output should be same with the padding one:
\begin{equation}
    \begin{aligned}
        X_\text{pad}&=[0]_i\oplus X\oplus [0]_j\\
        M_\text{pad}&=[0]_i\oplus M\oplus [0]_j\\
        f(X,M)&=f(X_\text{pad},M_\text{pad})[i:i+n]
    \end{aligned}
\end{equation}

Obviously, relative positions~\citep{relpos, t5, wang2020position, rotary} have this property instead of absolute ones~\citep{transformer, bert}. Even though absolute sinusoidal embedding has a similar property~\citep{transformer}: $PE_{pos+k}$ can be represented as a linear function of $PE_{pos}$, the addition operation in the initial word embedding messes the attention weight, where the spread form of $QK^T$ has 4 components whose geometric connection with position is unclear.

\subsection{Length Extrapolation}

As the cost of pre-training is getting bigger due to the larger model size and corpus, we do not hope to retrain a model just because of the longer length of downstream tasks.
A Transformer model with a suitable design should be capable of dealing with any input length.

First, learnable absolute position embedding~\citep{bert} is not able to extrapolate at all because it does not have any pre-defined position knowledge.
With the evaluation of perplexity on different length~\citep{alibi}, almost every position embedding's performance drops significantly~\citep{transformer, t5, rotary}.
Alibi~\citep{alibi} solves this problem by adding an exponential decay on the attention matrix, which lower the influence of out-of-distribution position like a soft sliding window. However, the absence of long-term dependency contributes to a performance drop compared with other relative strategies. Table~\ref{table:lm} shows that Alibi's perplexity is larger than \rope{} about 0.2 $\sim$ 0.3.

However, the extrapolation ability needs a systematic design where position embedding is a crucial but not only component.
With the proper attention map, the relative position can deal with long text, where the perplexity does not explode but does not decrease at the same time. The ideal situation is to use the long context in the right way, in that case, the model should perform better instead of saturation.




\section{A Length-Extrapolatable Transformer}
\label{sec:method}

We define attention resolution as the indicator of length extrapolation in Section~\ref{sec:resolution}.
Then we propose two ways to maximize the resolution metric, i.e., improve the length extrapolation of Transformers.
First, we introduce a relative position encoding method (Section~\ref{sec:xpos}) to explicitly maximize attention resolution.
Second, we propose to use blockwise causal masking (Section~\ref{sec:bca}) during inference for improved resolution.
The proposed architecture is named \textbf{L}ength-\textbf{Ex}trapolatable (\our{}) Transformer. 

\subsection{Attention Resolution}
\label{sec:resolution}

The monotonicity of attention scores is essential to represent distance in language models. We denote $s[n]$ as the score expectation when the distance of two tokens is $n$. We define \textit{attention resolution} $R(s)$ as a metric to evaluate attention's ability to recognize position:
\begin{equation}
R(s)=\sum_{i=0}^{N}\frac{e^{s[i]}(e^{s[i]}-e^{s[i+1]})}{(\sum_{i=0}^{N}e^{s[i]})^2}\\
\label{eq:reso}
\end{equation}

First, $s[i]>s[i+1]$ is preferred to ensure monotonicity. Besides, we implement $\softmax$ operation to simulate the attention probability. To mitigate the influence of long-tail distribution, the factor $e^{s[i]}$ is multiplied.
We can estimate $s[n]$ and $R(s)$ quantitatively when we design Transformers.


\subsection{Improve Resolution by Position Encoding}
\label{sec:xpos}

\citet{rotary} propose that by adding absolute position embedding on query and key, the attention matrix is actually encoded with relative position information. We use a similar but generalized strategy. First, a pseudo inner product is defined as $\langle x,y\rangle=\sum Re(x_i\cdot y_i^*)$, which is consistent with the exact inner product's definition when we map $\mathbb{C}^{d/2}\rightarrow \mathbb{R}^{d}$.
Formally, the encoding must satisfy:
\begin{equation}
\langle f_q(q,n+r),f_k(k, n)\rangle =\langle f_q(q, r), f_k(k, 0)\rangle 
\end{equation}

A simple solution is as follows:
\begin{equation}
    \begin{aligned}
        f_q(q,n)&=A_qqe^{\lambda n}\\
        f_k(k,n)&=A_kke^{-\lambda n}\\
    \end{aligned}
\end{equation}

The scaling factor $A_q, A_k$ is unnecessary because $q,k$ is obtained by a linear transformation. $\lambda =k+i\theta\in\mathbb{C}^{d/2}$ where $k,\theta\in\mathbb{R}^{d/2}$:
\begin{equation}
    \begin{aligned}
        f_q(q,n)&=qe^{\xi n+i\theta n}\\
        f_k(k,n)&=ke^{-\xi n-i\theta n}\\
    \end{aligned}
\end{equation}

If $\xi =0$, the form is the same as \rope{}~\citep{rotary}. Geometrically, the transformation provides a rotation on vectors. If the relative angle between $q$ and $k$ is larger, the inner product is smaller. However, the cosine value is not monotony if the rotating angle is large than $\pi$, which causes an unstable phenomenon that the expectation of the inner product oscillates dramatically with the growth of relative distance. Following the parameters~\citep{transformer, rotary} $\theta=\{\theta_i=10000^{-2i/d},i\in[0,1,...,d/2]\}$, we will calculate the expectation as follows. For generate models, we assume $\mathbb{E}(\angle q)\le\mathbb{E}(\angle k)$ to ensure the monotony:
\begin{equation}
    \begin{aligned}
        &\mathbb{E}[\langle qe^{m\xi +im\theta},ke^{n\xi+in\theta}\rangle]\\
        =&\sum_{i=0}^{d/2}\mathbb{E}[Re(\vq_i \vk_ie^{(m-n)\xi_i+i(m-n)\theta_i})]\\
        \le&\sum_{i=0}^{d/2}Re(\mathbb{E}[|\vq_i \vk_i|]e^{(m-n)\xi_i+i(m-n)\theta_i})\\
        \propto&\sum_{i=0}^{d/2}cos(m-n)\theta_ie^{(m-n)\xi_i}
    \end{aligned}
\end{equation}

The inference here is different from \citep{rotary} because of two reasons:
1) there is an additional assumption brought by language models;
2) the inequality scaling of \citep{rotary} is too strong to lose generality.
We calculate expectation instead of the upper bound.

Now we define a function to represent the property of relative position:
\begin{equation}
    g_\zeta[n]=\sum_{i=0}^{d/2} \cos n\theta_i \zeta_i^{n}
    \label{eq:exp}
\end{equation}

Stabilizing the curve of $g[n]$ is an intuitive way. Even though attention bias can achieve this goal, we do not hope additional position calculation. Instead, we can achieve this goal by selecting a good $\zeta$ to maximize $R(g_\zeta)$.

Obviously, the oscillation mainly comes from large $\theta_i$. Manually setting $\zeta$ can achieve this goal:
\begin{equation}
    \widetilde{\zeta_{i}}=\frac{i/(d/2)+\gamma}{1+\gamma}\in[0, 1]
\end{equation}
where $\widetilde{\zeta_{i}}$ becomes smaller when $\theta_i$ is larger.
In this way, we punish the oscillation of unstable dimensions and keep the distribution of stable ones.

Numerical optimization methods are tried to find optimal values for $\zeta$. However, the results rely on the initial value and lack control when the hidden dimension changes. Besides, the numerical precision should be considered because of fp16's range. Finally, we find a sub-optimal solution by manually setting $\gamma$ to both satisfy the resolution is recognizable ($R(g_\zeta)$ is partially optimized) and $\zeta_i^n$ can be represented by fp16 when $n$ is big (8192 in our setting). The optimized value $\hat{\zeta}$ will be used as the final value in \ourtrm{}.


\begin{figure}[t]
\centering
\includegraphics[width=\linewidth]{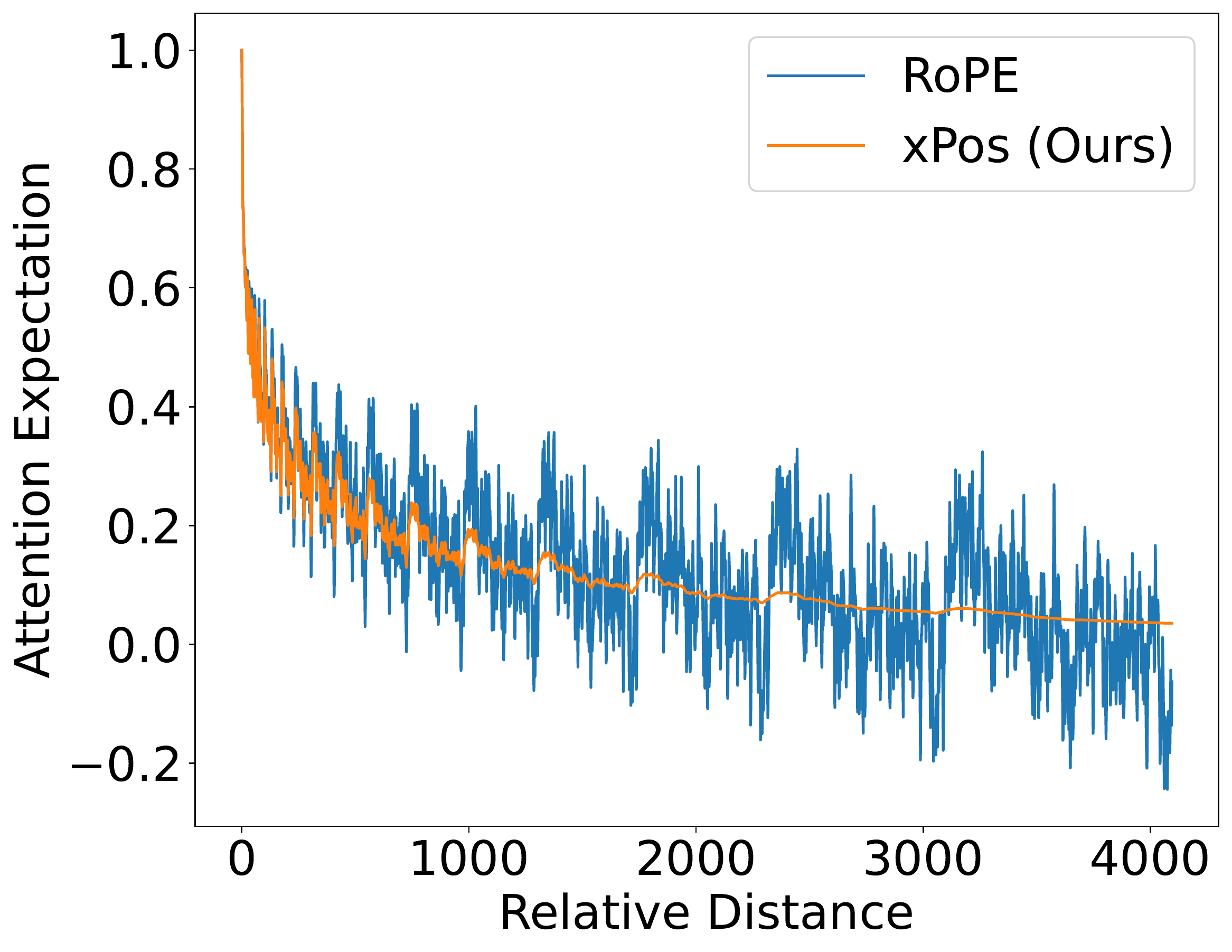}
\caption{
The long dependency curve of attention expectation. \rope{}'s dramatic oscillation confuses the attention resolution at long distances.
In contrast, \expo{} provides stable and accurate position modeling.
}
\label{fig:curve}
\end{figure}

The curves of $\zeta=\mathbf{1}, \hat{\zeta}$ are shown in Figure~\ref{fig:curve}.
The default rotary embedding contributes to a dramatic oscillation, especially in the large relative distance, which causes bad extrapolation performance and restricts the model's convergence speed. After adding a decay, the curve is almost stable, especially on long-term dependency. What's more, it does not hurt pure rotation's fitting ability because $\zeta_i^n \approx 1$ when $i$ is large or $n$ is small. In that way, short-term and long-term dependencies are divided continuously.

Finally, we have Extrapolatable Position Embedding (\expo{}):
\begin{equation}
\small
\begin{aligned}
f_q(q,n)&=
\begin{pmatrix}
q_1 \cos n\theta_1\hat{\zeta}_1^n-q_2 \sin n\theta_1\hat{\zeta}_1^n\\q_2 \cos n\theta_1\hat{\zeta}_1^n+q_1 \sin n\theta_1\hat{\zeta}_1^n\\ \vdots \\q_{n-1} \cos n\theta_{d/2}\hat{\zeta}_{d/2}^n-q_n \sin n\theta_{d/2}\hat{\zeta}_{d/2}^n\\q_n \cos n\theta_{d/2}\hat{\zeta}_{d/2}^n+q_{n-1} \sin n\theta_{d/2}\hat{\zeta}_{d/2}^n
\end{pmatrix}\\
f_k(k,n)&=
\begin{pmatrix}
k_1cosn\theta_1\hat{\zeta}_1^{-n}-k_2 \sin n\theta_1\hat{\zeta}_1^{-n}\\k_2 \cos n\theta_1\hat{\zeta}_1^{-n}+k_1 \sin n\theta_1\hat{\zeta}_1^{-n}\\ \vdots \\k_{n-1} \cos n\theta_{d/2}\hat{\zeta}_{d/2}^{-n}-k_n \sin n\theta_{d/2}\hat{\zeta}_{d/2}^{-n}\\k_n \cos n\theta_{d/2}\hat{\zeta}_{d/2}^{-n}+k_{n-1} \sin n\theta_{d/2}\hat{\zeta}_{d/2}^{-n}
\end{pmatrix}
\end{aligned}
\normalsize
\label{eq:sope}
\end{equation}

\begin{algorithm}[t]
\caption{Attention with \expo{}}
\label{alg:sope}
\textbf{def} $\mathrm{rot}(x)$:\\
~~\Return $[-x_1, x_0, -x_3, x_2, ...]$\\
\textbf{Initialization:}\\
$\theta_{i}=1/10000^{2i/d}$,~$\mathbf{\theta}\in\mathbb{R}^{d/2}$ \\
$\hat{\zeta}_i=(i/(d/2)+\gamma)/(1+\gamma),~\hat{\zeta}\in\mathbb{R}^{d/2}$\\
\KwIn{$Q,K,V\in\mathbb{R}^{h\times l\times d}, M\in\mathbb{R}^{d\times d}$}
$C_{mn} = \cos m\theta_n, C\in\mathbb{R}^{l\times d/2}$\\
$S_{mn} = \sin m\theta_n, S\in\mathbb{R}^{l\times d/2}$\\
$T_{mn} = \hat{\zeta}_n^m, T\in\mathbb{R}^{l\times d/2}$\\
$Q = (Q\times C + \mathrm{rot}(Q)\times S)\times T$ \\
$K = (K\times C + \mathrm{rot}(K)\times S)\times T^{-1}$ \\
$output = \softmax(\frac{QK^T}{\sqrt{d}}\cdot M)V$ \\
\Return $output$
\end{algorithm}

In the implementation, the transformation for key and value can be easily calculated by parallel addition and multiplication as shown in Algorithm~\ref{alg:sope}.

\subsection{Blockwise Causal Attention}
\label{sec:bca}

Another way to improve attention resolution (Section~\ref{sec:resolution}) is using windowed attention.
During inference, we use blockwise masking~\citep{transformerxl, bigbird, blockwise} for self-attention.
Notice that other window strategies, such as sliding window~\citep{sparsetransformer}, also work.
We use blockwise causal attention because it is cache-friendly and easy to implement.

\begin{figure}[t]
\centering
\includegraphics[width=0.98\linewidth]{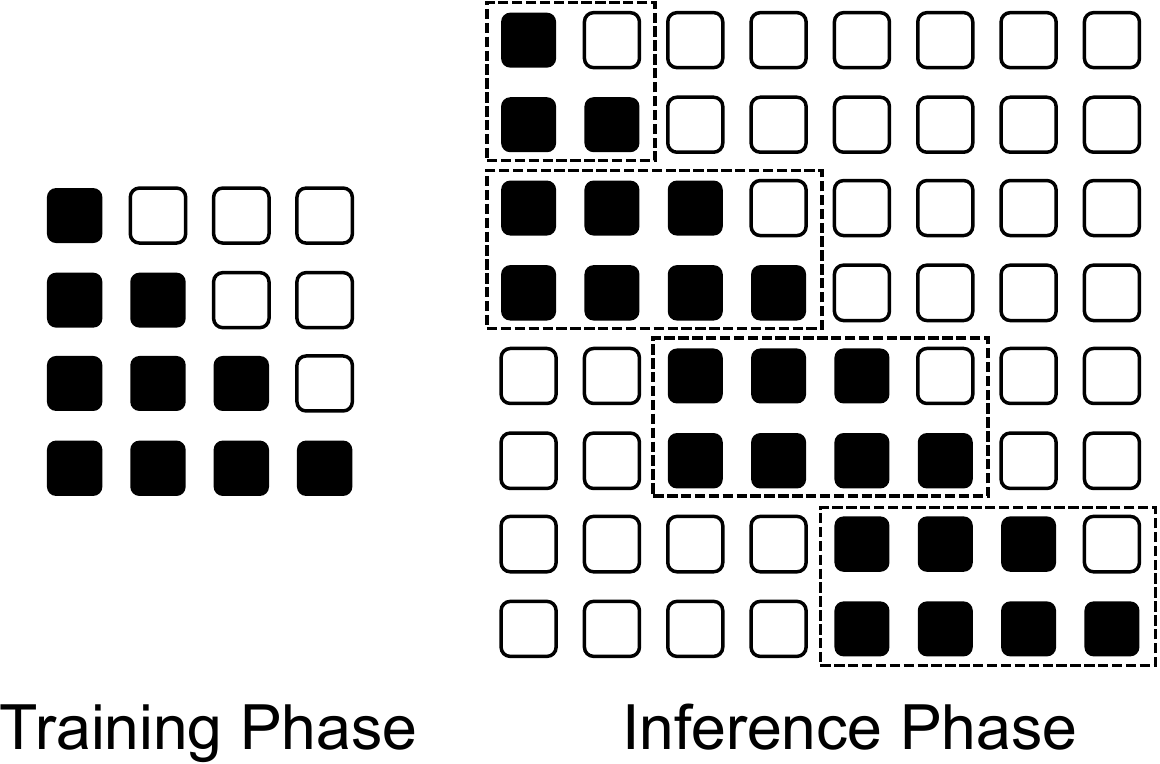}
\caption{
Our language model is trained on shorter texts in the same way as vanilla Transformers, i.e., using causal masking.
During inference, we use blockwise causal attention for longer sequences, which recurrently reuses the overlapped parts (i.e., key and value vectors).
}
\label{fig:block}
\end{figure}

As shown in Figure~\ref{fig:block}, if the pre-training length is $l$, we divide the query as blocks with $l/2$ length, and each query interacts with its own block and the last block. In this way, the context information can be delivered by the reuse of key and value. The window constraint helps models to encode longer input with improved resolution.

Different from training a long-sequence model with stop-gradient, we use vanilla attention in the training phase, because the pre-training corpus is not very long on average. However, during the inference phase, when dealing with long sequences, we directly implement BCA to help the model to be more position-recognizable.




\section{Experiments}
\label{sec:exp}

\subsection{Pre-training}

To fairly evaluate different Transformer variants, we pre-train the Transformer from scratch.
We use 1024 hidden dimension, 16 heads, and 24 layers, i.e., comparable to medium-size GPT-3~\citep{gpt3}.
The training corpus includes a subset of the Pile~\citep{pile}: Books3, OpenWebText2, Stack Exchange, PubMed Abstracts, Wikipedia, Gutenberg (PG-19), BookCorpus2, NIH ExPorter, and Pile-CC datasets.
The training procedure is implemented on 16$\times$V100 GPUs. Maximal length is 1024 for saving memory and extrapolation evaluation. The learning rate is $3\times10^{-4}$ and polynomial decay is used to adjust learning rate. The global batch size is 512 to follow GPT-3\citep{gpt3}, i.e., 0.5M token size. We use Adam~\citep{adam} optimizer with $\beta_1=0.9, \beta_2=0.98, \epsilon=10^{-6}$.
The code is based on TorchScale~\citep{torchscale}.

\subsection{Language Modeling}

\begin{table*}[t]
\centering
\begin{tabular}{@{}lccc|cc}
\toprule
\multirow{2}{*}{\bf Length} & \bf 256      & \bf 512      & \bf 1024     & \bf 2048     & \bf 4096     \\
& \multicolumn{3}{c}{Interpolation} & \multicolumn{2}{c}{Extrapolation} \\
\midrule
Transformer          & 46.34          & 36.39          & 29.94          & {\color{red} 132.63}          & {\color{red} 1283.79}         \\
Alibi             & 37.66          & 29.92           & 24.99          & 23.14          & {\color{red} 24.26}          \\
Roformer           & 38.09           & 30.38          & 25.52          & {\color{red} 73.6}          & {\color{red} 294.45}         \\
\ourtrm{} (Ours)   & \textbf{34.3} & \textbf{27.55} & \textbf{23.31} & \textbf{21.6} & \textbf{20.73} \\
\bottomrule
\end{tabular}
\caption{
Results of perplexity with different lengths.
The language models are trained with a length of 1024 and then evaluated on various lengths.
\our{} obtains better performance not only on shorter texts (i.e., interpolation) but also on longer texts (i.e., extrapolation).
The {\color{red} red} color indicates that the perplexity begins increasing compared with the shorter length.
\our{} is the only method that has lower perplexity along with increased evaluation length.
}
\label{table:lm}
\end{table*}

We first measure perplexity on arXiv, where the document length is usually larger than 6k, which can show the model's ability for long-dependency modeling. We care about the performance on different input lengths to evaluate the model's interpolation and extrapolation capability. For every document, we select its first 4k tokens and divide them into the target length to fairly compare the perplexity of different lengths. The results are shown in Table~\ref{table:lm}.

For interpolation capability, we analyze the results where the length is no more than 1024. All Transformers converge to similar perplexity. \expo{} have a stable advantage on others with 1$\sim$3 perplexity drop.

For lengths 2048 and 4096, we use BCA in all position embeddings, and the following ablation study will discuss the performance without that. \citet{alibi}'s experiment shows that most of the position strategies can't deal with input length longer than pre-training directly. In our experiment, with the improvement brought by BCA, \rope{} gets a better performance while Absolute still can't extrapolate. \expo{} shows a stable decrease when the sequence length increases, which satisfies the assumption that a longer context makes the prediction better. While others' perplexity increases when the input length is 4096.

Here, \expo{}'s advantage towards \rope{} is worth analyzing. With BCA, the position embedding does not extrapolate, so \rope{} also has the potential to encode long documents. However, with the forward layer by layer, the distribution of hidden states is different from pre-training. Then, the resolution matters to building a recurrent-similar encoding.

The experiment shows that \expo{} gets better performance on language modeling. With the stable advantage of any length, users can input any sentence freely without the concern of position. Besides, results also indicate that is not essential to build an explicit decay on the attention matrix, Instead, a proper design for an attention mask is actually better to deal with long-context tasks.

\subsection{Measuring Resolution}


\begin{table}[t]
\centering
\begin{tabular}{lcc}
\toprule
\multirow{2}{*}{\bf Length}          & \bf 1024 & \bf 2048 \\
& Interpolation & Extrapolation \\
\midrule
Transformer & 0.87 & 0.28 \\
Alibi               & 0.81 & 0.88 \\
Roformer            & 0.91 & 0.08 \\
\our{} (Ours)     & \textbf{0.98}  & \textbf{1.08} \\
~~~$-$ BCA              & \textbf{0.98} & 0.54 \\
\bottomrule
\end{tabular}
\caption{Results of resolution with different Transformer variants.
Higher resolution indicates that the architecture tends to better distinguish context tokens.
``BCA'' is short for blockwise causal attention.
}
\label{table:reso}
\end{table}

In the previous section, we claim that resolution is a crucial index for building an effective Transformer. To verify the claim, we evaluate the resolution of different Transformer variants empirically. Equation~\ref{eq:exp} estimates the expectation of attention score for \our. Denote attention score of query $i$ and key $j$ (before $\softmax$) as $e_{ij}$, the expectation of $s[n]$ is as follows:
\begin{equation}
    \hat s[n]=\mathbb{E}[s[n]]=\frac{1}{N-n}\mathbb{E}[\sum_{i=n}^{N-1}e_{i(i-n)}]
    \label{eq:emexp}
\end{equation}

The resolution can be calculated by combining Equation~\ref{eq:reso} and \ref{eq:emexp}.  The final expectation is the average of different input text. Resolution is calculated in every layer, and the average resolution is shown in Table~\ref{table:reso}.  The results show that \expo{} makes the position more recognizable in training length (1024). For Alibi~\cite{alibi}, the stable resolution comes from explicit decay, but it prevents the model from learning position dependency itself. Besides, we run an ablation on BCA. In length 2048, we measure the resolution with/without block. The result supports that BCA helps model distinguish positions better.

\subsection{Ablation Studies}

\subsubsection{Rotation Computation}

\begin{table}[t]
\centering
\begin{tabular}{ll}
\toprule
\bf Methods & \bf Perplexity   \\
\midrule
\rope{} & 17.74 \\
\expo{} (Ours) & \textbf{17.54} \\
~~$-$ Rotation & 33.68 \\
\bottomrule
\end{tabular}
\caption{Ablation results on the validation set show that rotation of \expo{} is necessary for strong performance.}
\label{table:ablation_rot}
\end{table}

In this part, we discuss the necessity of the combination of vector rotation and exponential decay. \expo{} without rotation means Equation~\ref{eq:sope} degenerates to $\theta_i=0$:
\begin{equation}
\dot{f}_q(q,n)=
\begin{pmatrix}
q_1\hat{\zeta}_1^n\\q_2\hat{\zeta}_1^n\\ \vdots \\q_{n-1}\hat{\zeta}_{d/2}^n\\q_n\hat{\zeta}_{d/2}^n
\end{pmatrix}
\dot{f}_k(k,n)=
\begin{pmatrix}
k_1\hat{\zeta}_1^{-n}\\k_2\hat{\zeta}_1^{-n}\\ \vdots \\k_{n-1}\hat{\zeta}_{d/2}^{-n}\\k_n\hat{\zeta}_{d/2}^{-n}
\end{pmatrix} \nonumber
\end{equation}

After pre-training, we test the perplexity on the valid split of training corpus with 1k length. The result in Table~\ref{table:ablation_rot} shows that simple scaling operation can't perform as well as \our. Therefore, the combination of rotation and decay means the combination of in-distribution and out-of-distribution ability.

\subsubsection{Blockwise Causal Attention}

\begin{table}[t]
\centering
\begin{tabular}{lll}
\toprule
\multirow{2}{*}{\bf Methods}    & \bf 2048  & \bf 4096  \\
& \multicolumn{2}{c}{Extrapolation} \\
\midrule
\rope{}          & 73.6 & 294.45 \\
\rope{} + BCA          & 25.57 & 25.65 \\
Alibi          & 23.14 & 24.26 \\
Alibi + BCA           & 24.6 & 25.37 \\
\expo{} (Ours) & 22.56 & 28.43 \\
\expo{} + BCA (Ours)           & \textbf{21.6} & \textbf{20.73} \\ \bottomrule
\end{tabular}
\caption{
Results of perplexity on arXiv dataset.
``BCA'' is short for blockwise causal attention.
}
\label{table:ablation_block}
\end{table}

To fairly compare different methods, we run the evaluation using different position embeddings (i.e., Alibi, \rope{}, and \expo{}) with or without blockwise causal attention.
The results are shown in Table~\ref{table:ablation_block}.

First, Blockwise Causal Attention works for \rope{} whose perplexity will explode without that. Alibi performs well without windowed attention because its ``soft window'' is broader than a hard block window.
\expo{}'s perplexity without BCA increases by about 1 in 2048, and 8 in 4096.
However, with its high resolution, \expo{} can recognize position with BCA's constraint.

\section{Related Work}

\subsection{Long-Sequence Transformers}

Long-sequence Transformers aim to solve two key problems.
First, the computation or memory consumption is not efficient enough for long sequences.
Second, there is a trade-off between performance and efficiency.

One popular solution~\cite{linformer, linear-transformer, performer} is linear attention, i.e., using a kernel-based or low-rank approximation to replace vanilla attention. The methods typically target efficiency while underperforming vanilla Transformers for regular length.
Another strand is sparse attention~\cite{sparsetransformer,longformer,bigbird,blockwise}, which usually leverages structured sparsity to reduce computation.
For causal sequence modeling, the recurrent-style designs~\cite{transformerxl,blockrecurrenttrm,mega} are also competitive.

In comparison, we focus on the extrapolation issue~\cite{alibi} for language modeling, i.e., training on short texts while evaluating long texts.
The training process is kept the same as vanilla Transformers, i.e., training on short sequences, and using dense attention computation.
The capability of long-sequence modeling is given for free during inference.
So the training efficiency (which is typically expensive for large-scale language models) is not affected compared with previous work.
Moreover, the performance on regular length is perfectly retained, without trade-offs for long-sequence modeling.



\subsection{Position Modeling}

\subsubsection{Absolute Position Embedding}

Absolute sinusoidal position embedding is proposed by \citet{transformer}. For each dimension, different frequencies are encoded from $2\pi$ to $10000\times 2\pi$:
\begin{equation}
\begin{aligned}
\text{PE}_{(pos, 2i)}=\cos(pos/10000^{2i/d_\text{model}})\\
\text{PE}_{(pos, 2i+1)}=\sin(pos/10000^{2i/d_\text{model}})
\end{aligned}
\end{equation}
where $\text{PE}_{pos+k}$ is represented as a linear function of $\text{PE}_{pos}$ to restore a relative-position property.


\subsubsection{Relative Position Embedding}

\citet{relpos} propose relative position embedding as an alternative approach.
Denote $e_{ij}$ as attention weight, $\alpha_{ij}=\softmax(e_{ij})$, $o_i$ as output, we have:
\begin{equation}
    \begin{aligned}
        e_{ij}=\frac{\vq_i\cdot \vk_j}{\sqrt{d}}&\Longrightarrow \frac{\vq_i\cdot(\vk_j+ \va_{ij}^K)}{\sqrt{d}}\\
        o_i=\sum_j \alpha_{ij}\vv_j&\Longrightarrow \sum_j \alpha_{ij}(\vv_j+ \va_{ij}^V)
    \end{aligned}
\end{equation}
where $\va_{ij}^K=\omega_{\mathrm{clip}(i-j, k)}^K, \va_{ij}^V=\omega_{\mathrm{clip}(i-j, k)}^V$, and $\omega^K$ and $\omega^V$ are learnable parameters.
The clipping strategy helps length generalization but cannot distinguish the positions that are larger than $k$.
\citet{xlnet} and \citet{deberta} further reparameterize the relative position vectors for better performance.
%
%
T5~\citep{t5} uses a simpler strategy to encode relative position:
\begin{equation}
    e_{ij} = \frac{\vq_i\cdot \vk_j}{\sqrt{d}} + a_{\mathrm{bucket}(i-j)}
\end{equation}
where log-bucket scalars are added to attention scores.
%
Recently, pre-defined position embedding is brought back by \rope{}~\citep{rotary}.
Alibi~\citep{alibi} proposes to explicitly build an exponential decay on the attention matrix, which contributes to length extrapolation:
\begin{equation}
e_{ij} = \frac{\vq_i\cdot \vk_j}{\sqrt{d}} - \mathrm{m}(i-j),~~~~\mathrm{m}(\cdot) > 0
\end{equation}
where the values of $\mathrm{m}(\cdot)$ are manually defined.
However, Alibi~\citep{alibi}'s performance tends to be inferior to \rope{} for the context whose length is shorter than the pre-training length.
In this work, we propose a theoretically derived relative position embedding \expo{} that optimizes the attention resolution between tokens.
The \expo{} method not only has the nice property of length extrapolation but also achieves strong performance.



\section*{Limitations}

In this work, we focus on causal language modeling. It needs additional efforts to integrate the proposed methods into bidirectional attention, such as masked language modeling~\citep{bert}.
Moreover, \expo{} introduces about 6\% inference cost compared with absolute position embeddings, although it accelerates training convergence.



\bibliography{anthology,relpos}
\bibliographystyle{acl_natbib}






\end{document}